\documentclass[conference]{IEEEtran}
\IEEEoverridecommandlockouts
\usepackage[left=1.62cm,right=1.62cm,top=1.9cm]{geometry}
\usepackage{cite}
\usepackage{amsmath,amssymb,amsfonts}
\usepackage{algorithmic}
\usepackage{graphicx}
\usepackage{textcomp}
\usepackage{xcolor}
\usepackage{xurl}
\def\BibTeX{{\rm B\kern-.05em{\sc i\kern-.025em b}\kern-.08em
    T\kern-.1667em\lower.7ex\hbox{E}\kern-.125emX}}
\begin{document}

\title{Identifying Risk Patterns in Brazilian Police Reports Preceding Femicides: A Long Short Term Memory (LSTM) Based Analysis\\

}

\author{\IEEEauthorblockN{Vinicius Lima}
\IEEEauthorblockA{\textit{Polytechnic Institute} \\
\textit{Purdue University}\\
West Lafayette, USA \\
vlima@purdue.edu}
\and
\IEEEauthorblockN{Jacque Almeida de Oliveira}
\IEEEauthorblockA{\textit{Criminalistics Institute} \\
\textit{Civil Police of the Federal District}\\
Brasilia, Brazil \\
jaqueline.oliveira@pcdf.df.gov.br}

}

\maketitle

\begin{abstract}
Femicide refers to the killing of a female victim, often perpetrated by an intimate partner or family member, and is also associated with gender-based violence. Studies have shown that there is a pattern of escalating violence leading up to these killings, highlighting the potential for prevention if the level of danger to the victim can be assessed. Machine learning offers a promising approach to address this challenge by predicting risk levels based on textual descriptions of the violence. In this study, we employed the Long Short Term Memory (LSTM) technique to identify patterns of behavior in Brazilian police reports preceding femicides. Our first objective was to classify the content of these reports as indicating either a lower or higher risk of the victim being murdered, achieving an accuracy of 66\%. In the second approach, we developed a model to predict the next action a victim might experience within a sequence of patterned events. Both approaches contribute to the understanding and assessment of the risks associated with domestic violence, providing authorities with valuable insights to protect women and prevent situations from escalating.
\end{abstract}


\section{Introduction}

Femicide or Feminicide is a hate crime that is broadly defined as the "intentional killing of women or girls because they are female", however, definitions of the term may vary depending on the cultural context \cite{wikipedia_femicide_2022, grzyb2018femicide}. In this paper, femicide will be considered a crime related to domestic violence against female individuals according to the Brazilian law definition \cite{avila2018facing}. The term gained notoriety after researchers and activists studied murder cases separated by gender, finding that murders of female victims have a high frequency of cases related to domestic violence, where the murderer is usually a family member or partner of the victim. In contrast, no specific motivations behind killings were found among the male population \cite{vives2016expert}. A 2017 study conducted by the United Nations (UN) demonstrated that when looking at the 20\% of female murders that occur around the world, 82\% were killed by an ex-partner \cite{uk_article}.
 
In Brazil, the situation is no different. In a recent study, Brazilian researchers demonstrated an increase of 4.8\% in femicide cases from 2004 to 2015 \cite{feminicide_brazil_article}. Since March 2015, the country has had specific punishment for femicide cases. While “regular” homicide authors can go to prison for 6 to 20 years, in femicide cases, the law regulates a minimum of 12 years and the maximum time in prison in Brazil, which is 30 years \cite{feminicide_law_brazil_article}. Although having a strict policy, the law did not stop the rise in cases over the past few years, and Brazil is still considered one of the top countries where women are killed in domestic violence \cite{waiselfisz2015mapa}. The scenario is worse when considering specifically the black population \cite{feminicide_brazil_article}. According to the World Health Organization (WHO), the country registered one rape every ten minutes, in 2021 \cite{feminicide_brazil_article}. Furthermore, according to data from the Brazilian Public Security Forum, in 2021 there were a total of 1,319 femicides in the country, that is, on average, a woman was killed every seven hours \cite{FBSP}.

Sociologists and Criminologists have demonstrated that before a man kills his female partner (or family member) previous violence had already been committed, and there is an increase in abuse and violence along the toxic relationship \cite{monckton_smith_intimate_2020}. It is not uncommon to see police incident reports of other physical and psychological violence cases before the final murder act. This highlights that some female murders could be avoided or, at the very least, safe resources could be provided to the victims beforehand. Therefore, the development of a risk assessment tool for domestic violence against women becomes imperative in order to prevent situations from deteriorating further.

The National Polytechnic Institute (Instituto Politécnico Nacional – IPN) associated with the Mexican Government has created a tool to help women understand the level of risk they may face in a relationship \cite{ipn_portal}. The tool, known as the "Violentómetro" or "Violent-o-meter", is a scale from 0 to 30, where zero represents a non-violent situation, and 30 represents the ultimate murder or femicide. In between, there is a progression of violent actions observed in the aggressor. This measuring scale has been adapted to other languages and is commonly used by activists to present the potential harm that domestic violence victims may face. See Fig. 1.

\begin{figure*}
    \centering
    \includegraphics[width=0.8\textwidth]{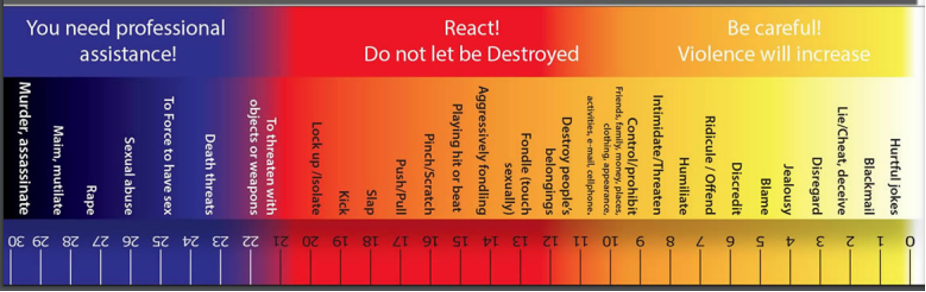}
    \caption{Violent-o-meter scale created by IPN \cite{ipn_portal}.}
    \label{fig:img1}
\end{figure*}

We aim to leverage machine learning, specifically the Long Short Term Memory (LSTM) technique \cite{hochreiter1997long}, to analyze police reports on domestic violence. Our objective is to assess the risk levels associated with escalating violence and potential femicide incidents. This analysis will provide valuable insights for law enforcement and social work authorities, enabling them to comprehend the circumstances surrounding female victims in reported cases. By understanding these risk levels, appropriate safety resources can be provided to prevent further violence and mitigate the potential for harm \cite{hoyle2008will}. Our developed model will be evaluated using textual data extracted from police reports in the Federal District of Brazil, specifically focusing on instances leading up to femicide cases.

We acknowledge that our proposed technique is not intended to replace existing methods for assessing the risk of domestic violence. Instead, we view it as a supplementary tool to aid police officers in evaluating victims' situations, particularly when they face challenges that hinder their ability to do so effectively. These challenges may include insufficient training in dealing with gender-based crimes \cite{eigenberg_confronting_2012}, limited access to historical case records \cite{jordan_beyond_2004}, and inadequate inter-agency coordination, such as when victims register complaints across different police stations. Furthermore, our approach offers the advantage of minimizing the potential for human bias to underestimate the severity of domestic violence \cite{skeem_gender_2005}. 

\section{Related Work}

Numerous studies have explored various approaches to risk assessment for femicide and violence against women \cite{dutton2000review, van2019predicting, myhill2019golden}. While much of the research in this area falls under the purview of Sociology and related fields, some researchers in Computer Science have also sought to address this issue. In particular, we are interested in works that leverage machine learning to predict outcomes of domestic violence or provide risk assessment. Since we ought to adopt an interdisciplinary perspective in our research, we will provide a brief overview of previous works that are relevant to our theme.

Smith, J. conducted a study of 372 intimate partner femicides, with the aim of identifying patterns of behavior and sequential markers that could help in risk assessment \cite{monckton_smith_intimate_2020}. The study revealed that certain markers, such as abuse, stalking, possessiveness, control, monitoring, violence, sexual violence, isolation, threats to kill, threats to commit suicide, stalking, separation, and escalating control or violence, were present in most of the cases studied. By identifying eight sequential stages leading up to the killing, the author demonstrated that understanding the phase the victim is in can be a critical factor in protecting these women.

A significant contribution to the field was made by Campbell et al., who conducted a seminal study that involved updating the Danger Assessment tool \cite{campbell1986nursing}, originally developed by the first author, to assess the potential for lethality or near lethality in cases of intimate partner violence \cite{campbell2009danger}. The revised version of the assessment incorporated modifications to the report sheet, providing guidance for authorities when interviewing victims of abuse. Notably, the study conducted in 11 cities demonstrated an impressive 90\% accuracy rate when testing the revised version \cite{campbell2004helping}. 

Risk assessment is a challenging and subjective task, as evidenced by Skeem, J.'s research, which demonstrated that mental health professionals face difficulties making accurate risk diagnoses in cases of domestic violence \cite{skeem_gender_2005, campbell2004helping}. Interestingly, this challenge affects both male and female professionals, and it is explained by the previously mentioned statistics that suggest women are subjected to lower rates of violence, leading to an underestimation of their situations. Additionally, crimes against women often occur in private and domestic settings, making them less visible and less likely to be reported by third-party individuals who may be hesitant to interfere or report.

Other studies used machine learning techniques to support risk assessment in domestic violence and gender-related crimes. González-Prieto et al. utilized the Nearest Centroid (NC) algorithm to classify police questionnaires completed by female victims in cases of domestic violence, with the aim of predicting the likelihood of recidivism in three categories: zero, low, and high \cite{gonzálezprieto2021machine}. The dataset used in this study was obtained from the Spanish Police's VioGen system, and the victims were asked a series of questions about the abuse they had experienced, with responses provided on a Likert scale. The system then generates a risk level of recidivism based on the answers provided. The authors compared the performance of the NC algorithm to that of the traditional VioGen system and found that NC achieved an f1 score of 14\% for the high-risk classification, compared to 10\% for the traditional method. They also proposed a hybrid solution that combines the algorithm with the police's traditional system.

Sharifirad and Matwin developed a study in which they applied NLP techniques to classify Twitter messages containing potential sexist content \cite{sharifirad2019tweet}. They utilized LSTM to predict tweets in five categories: Indirect Harassment, Information Threat, Sexual Harassment, Physical Harassment, and non-sexist content. LSTM performed better than previous algorithms, achieving an accuracy rate of around 91\%.

Similarly, Subramani et al. employed Deep Learning techniques to detect victims of domestic violence through analysis of Facebook posts \cite{subramani2018domestic}. They explored various models and found that the combination of Gated Recurrent Network (GRU) \cite{DBLP:journals/corr/ChoMGBSB14} and Glove embeddings \cite{pennington2014glove} outperformed other approaches, accurately classifying the data. Such assessments play a crucial role in providing real-time support to victims, allowing for timely intervention and assistance. 

D'Ignazio et al. employed machine learning to classify media and news articles potentially related to femicide \cite{dignazio2020feminicide}. Their objective was to create a database of information on the subject matter. To accomplish this, they developed an algorithm using Naïve Bayes, which accurately predicted 81\% of the texts.

Arian Yallico, T. and Fabian, J. conducted a study using data from six sources, including news media, social media, YouTube transcripts, campaigns, and anonymous report lines. Their objective was to train a classification model that could detect psychological violence in five categories: "Low Risk," "Emotional Blackmail," "Jealousy/Justification," "Insults/Humiliations," and "Threats/Possessiveness" \cite{yallico2022automatic}. The authors tested different algorithms and found that the unidirectional LSTM outperformed the others, achieving an accuracy of 93\%.

\section{Objectives}

Building on the reviewed literature, LSTM has emerged as a promising approach for classifying written content concerning domestic violence. Motivated by this, our objective is to leverage this technique to create a risk assessment tool that analyzes police reports encompassing cases starting from domestic violence and culminating in femicide incidents. By developing this tool, our project aims to equip law enforcement agencies with valuable knowledge resource that could be used to support public policies focused on safeguarding women in perilous situations.

Our approach delves into the advantages of utilizing machine learning within the context of natural language as a means to address classification and prediction challenges. Recognizing that domestic violence often exhibits a pattern of sequential behavior, which is also reflected in police reports, we divided our work into two distinct objectives. Firstly, we focused on accurately classifying the reports based on two levels of risk. Secondly, we aimed to predict the subsequent behavior anticipated within a sequence of actions. By adopting this approach, we aim to introduce an objective perspective that complements subjective analysis, thereby enhancing our understanding of the levels of danger depicted in pre-femicide data.

\section{Methodology}

\subsection{Dataset}

The dataset that is being used consists of police reports from the Civil Police of the Federal District, in Brazil \cite{noauthor_civil_2022}. After a domestic violence crime is registered, victims, suspects, and witnesses (if any) are invited to relate their versions of what happened. Based on their testimonies, police officers write brief descriptions in police reports. We will focus on this small history text to build our analysis. To understand the pattern of behavior in crimes against women, we gathered data on femicide cases, where some previous domestic violence was also found. Data were anonymized and provided by the Forensic Science Institution “Fundação de Peritos em Criminalística Ilaraine Acácio Arce”, from a previous internal study conducted by the Crime Against Life Sector in the Civil Police of the Federal District. From 2017 to 2021, the dataset consists of 142 femicide cases, of which 39 presented previous domestic violence crimes reported, corresponding to a total of 162 police reports. Fig. 2 presents how the cases are distributed throughout these five years.

\begin{figure}[htbp]
\centerline{\includegraphics[scale=0.8]{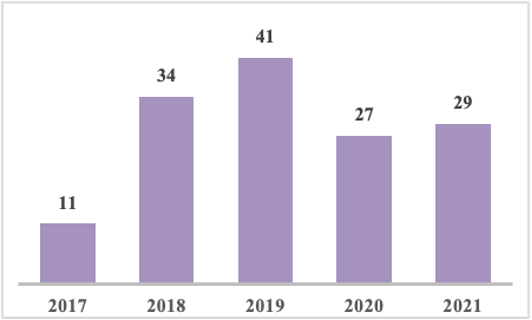}}
\caption{Femicide cases in the Federal District of Brazil.}
\label{fig}
\end{figure}

Although it is not the scope of this research, there is evidence that femicide police registration is underestimated due to several reasons \cite{trautman2007intimate, jordan_beyond_2004}. Moreover, as mentioned before, a significant number of victims deny reporting their abuse \cite{skeem_gender_2005, campbell2004helping}. Previous work from a different region in Brazil demonstrated that some femicides are neglected and some are registered as homicides \cite{margarites_feminicides_2017}. This highlights the significance of drawing attention to this problem and serves as an additional motivation to raise awareness within the community.
 
Through a word cloud visualization constructed from the 162 reports (Fig. 3), we can see the main words presented (written in Portuguese). When translated, we can see that among the principal words are “Victim”, “Home”, “Aggression”, and “Protective Measures”, providing a picture of the domestic violence scenario.

\begin{figure}[htbp]
\centerline{\includegraphics[scale=0.8]{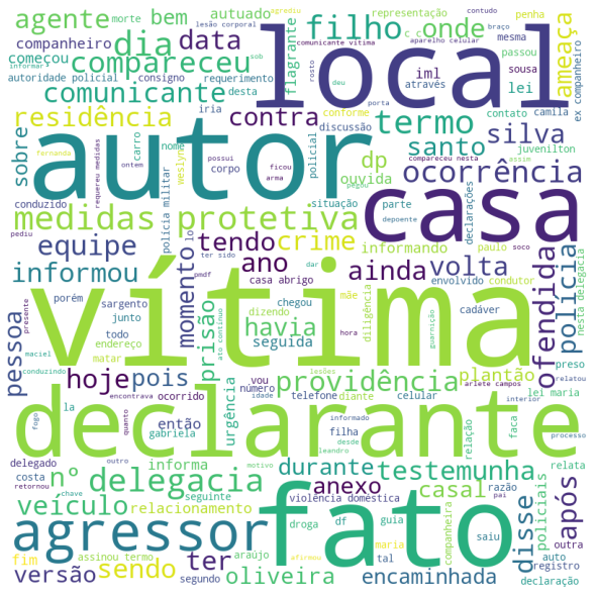}} 
\caption{Word cloud (in Portuguese) resulted from the 162 reports.}
\label{fig}
\end{figure}

\subsection{Preprocessing and Modeling}

The project encompassed two approaches aimed at exploring how NLP can extract pattern information from domestic violence reports. The first approach involved a straightforward classification of the reports into two categories: low-risk and high-risk. By analyzing the dates of the murder and previous reports, we were able to calculate the time interval between each report and the day of the femicide. This provided us with an understanding of the level of risk the victim faced in terms of being murdered by her partner or family member. Some cases involved multiple reports leading up to the final killing, while others had only one recorded incident. On average, we observed that there were 1,336 days preceding the femicide event. For the purpose of this study, we defined a threshold of one year (365 days) to determine the higher risk category, considering anything beyond that threshold as a lower risk category (though medium risk may be a more suitable term). These thresholds can be adjusted based on the perspective of the authorities involved. Following this approach, we identified 39 reports categorized as high risk and 79 reports classified as low risk (excluding reports specifically describing the final murder, and focusing solely on reports preceding the femicide).

\begin{table}[htbp]
\caption{Reports' labels}
\begin{center}
\begin{tabular}{|c|c|c|}
\hline
\textbf{Higher risk level} & {Less than one year to femicide}& {39 cases}\\
\hline
\textbf{Lower risk level} & {More than one year to femicide} & {79 cases}\\
\hline
\end{tabular}
\label{tab1}
\end{center}
\end{table}

Preprocessing the data is an essential step before feeding it into the algorithm. This involved removing symbols from the text, converting all letters to lowercase, and eliminating Portuguese stopwords, which are non-informative words like prepositions and pronouns. The text was then tokenized, and the words were converted into numerical vectors in a vector space. To ensure consistent input lengths, padding was applied to each vector. We labeled each report into 0, for higher risk (less than one year preceding the femicide), or 1, for lower risk (more than one year before the femicide). 

Our model was trained using LSTM on 70\% of the dataset and tested on the remaining 30\%. We employed 100 cells, a batch size of 64, Adam as the optimizer, a dropout rate of 0.2, and 30 epochs for training.

The second approach involved utilizing LSTM to predict the next action on a domestic violence scale. Similar to traditional methods used for predicting the next word in a sentence, our objective here was to anticipate the expected behavior if no further prevention measures were taken. To achieve this, features or key actions and behaviors were extracted from each report and arranged in a sequential list, which served as input to the model. The selection of features was based on relevant literature and identified patterns in offender-victim relationships. For example, from a specific report, keywords such as "yelled," "punch," and "threat" were extracted in that specific order. The aim was to create a sequence of actions that revealed the behavioral pattern. This approach can be considered an additional form of risk assessment, as the results can provide authorities with insights into the level of danger the victim faces and an understanding of what the woman may experience next. These insights can be valuable in formulating appropriate safety measures.

The extraction of keywords was done manually. Every report was read fully, and special attention was given to the victim's testimony, when present. The supporting knowledge base to extract features was the markers provided by Smith, J. \cite{monckton_smith_intimate_2020} and the "Violent-o-meter" scale \cite{ipn_portal}. In summary, the methodology used was the following:
\begin{enumerate}
\item Find and highlight keywords presented in the markers and/or the "Violent-o-meter", considering the word stem in order to catch all word derivations (for instance, verb conjugation). 
\item Read the full reports and extract implicit keywords. In some cases, a keyword may not be explicit in the text, only its semantics. For instance, in the sentence "If you break up with me, I will kill you," we observed an implicit death threat. The co-author of this paper, who is a police officer specializing in femicide crimes, helped in this inference process.
\item Set the words in a sequence. For this part of the project, we only used the cases where the murderer was also the author of previous violent crimes. Following this analysis, we got 22 sequences of reports. Based on the date registered, we ordered a sequence of actions from the victim's very first report until the final femicide report. 
\end{enumerate}

Table 2 shows the key action/behavior and the number of times they appeared in the reports. It is important to mention that the content seen in the police reports was diverse with respect to details. A few provided just simple broad overviews of the charges and technical law descriptions, but the majority contained detailed narratives, such as the victims’ quoted dialogues. Also, some reports presented not only the victim’s version but also the author or the witnesses. In these cases, we only considered the victims’ perspective to extract the key characteristics.

\begin{table}[htbp]
\caption{List of features extracted for sequence analysis. }
\begin{center}
\begin{tabular}{|c|c|}
\hline
\textbf{Extracted Feature} & \textbf{Frequency}\\
\hline
{Verbal Offense} & {29}\\
\hline
{Physical Violence} & {21}\\
\hline
{Death Threat} & {15}\\
\hline
{Discussion} &	{14}\\
\hline
{Threat} &	{13}\\
\hline
{Jealousy} &	{8}\\
\hline
{Physical Fight} &	{7}\\
\hline
{Punches} &	{7}\\
\hline
{Physical Threat} &	{4}\\
\hline
{Object Damage} &	{4}\\
\hline
{Break Deny} &	{4}\\
\hline
{Hair Pull} &	{3}\\
\hline
{Kick} &	{3}\\
\hline
{Stalk} &	{3}\\
\hline
{Biting} &	{3}\\
\hline
{Strangling} &	{3}\\
\hline
{Slap} &	{2}\\
\hline
{Push} &	{2}\\
\hline
{Sexual Harassment} &	{2}\\
\hline
{Residence Invasion} &	{2}\\
\hline
{Possessive Control} &	{2}\\
\hline
{Relationship Persistence} &	{1}\\
\hline
{Rape} &	{1}\\
\hline
\end{tabular}
\label{tab1}
\end{center}
\end{table}

In this second approach, we also preprocess the data to feed into the algorithm. However, since we already have the main keywords, we did not need to remove symbols or lowercase the words. We moved forward to tokenize each word or term and then added zero padding to each sequence. Finally, we created an LSTM architecture with 100 cells, Adam as the optimizer, 0.1 dropout, and 50 epochs. In both analyses, we used Python (along with Keras and TensorFlow libraries) to build the model.

\section{Results and Discussion}
\subsection{LSTM for identification of higher or lower risk.}

Our primary objective can be likened to performing sentiment analysis on the text data. After conducting tests, we achieved a reasonable accuracy of 66\%, taking into account the subjective nature of the task and the strong presence of pattern-sequential behavior.

It is worth noting that the analysis encompassed the entire content of the report descriptions. While the majority of cases involved the same perpetrator as previous domestic violence crimes, there were also instances where the murderer in the femicide case was unknown or unrelated to previous reports. Nonetheless, employing machine evaluation of risk presents an interesting approach to supplement decision-making and allocate resources for the protection of these women victims.

A criminal report classified as high-risk, despite a 66\% accuracy, should serve as a catalyst for authorities to take action before a more severe outcome occurs, irrespective of the identity of the perpetrator. The aim is to prioritize the safety of potential victims and mitigate risks associated with domestic violence situations.

\subsection{LSTM for sequence behavior prediction.}
In our second approach, we employed LSTM to perform a sort of "next word" analysis. The objective was to extract key features, in the form of keywords or terms, from the police reports and establish a sequential pattern that the algorithm could utilize to predict the subsequent action or key feature.

Once LSTM was trained with the defined sequences, we were able to observe outputs given a sequence of words. For instance, the sequence "Discussion, Verbal Offense, Physical Violence" resulted in the output "Verbal Offense," while the sequence "Punches, Verbal Offense, Physical Violence, Physical Violence, Death Threat" yielded the output "Feminicide." Refer to Table III for examples of this next behavior prediction approach.

 \begin{table}[htbp]
\caption{Examples of LSTM predicting next action abuse.}
\begin{center}
\begin{tabular}{|c|}
\hline
Discussion \textrightarrow Verbal Offense \textrightarrow Physical Violence \textrightarrow \textbf{Verbal Offense}\\
\hline
Punches \textrightarrow Physical Violence \textrightarrow Physical Fight  \textrightarrow Verbal Offense \textrightarrow\\
\textbf{Death Threat}\\
\hline
Punches \textrightarrow Verbal Offense \textrightarrow Physical Violence \textrightarrow Physical Violence \textrightarrow \\
Death Threat \textrightarrow  \textbf{Femicide}\\
\hline
Object Damage \textrightarrow Physical Violence \textrightarrow Verbal Offense  \textrightarrow Rape \textrightarrow \\ 
Sexual Harassment \textrightarrow Verbal Offense \textrightarrow Death Threat \textrightarrow \\
Verbal Offense \textrightarrow Death Threat \textrightarrow \textbf{Femicide}\\
\hline

\hline
\end{tabular}
\label{tab1}
\end{center}
\end{table}

During the experiments, we made an important observation: some broad definitions needed to be broken down into more specific words. For example, the term "Physical Aggression" (or simply "aggression") appeared in nearly all the reports. When using this broad term, the algorithm consistently produced the same output value. This also applied to the term "Threat." To provide the algorithm with a greater understanding of the data nuances, we disaggregated "Physical Aggression" into more detailed words such as "Punches," "Slaps," or "Strangle" whenever such details were present. Similarly, "Threat" was further broken down into "Physical Threat" and "Death Threat."

The results once again highlighted the potential usefulness of this tool in determining the level of risk for victims of domestic violence. The analysis can offer victims and authorities a more realistic perspective on the potential harm that may occur if no action is taken to protect the victim. We argue that this method can provide a more impactful response to the victim than a simple indication of the risk of criminal recidivism. Additionally, when examining individual reports in isolation, although some patterns may be discernible, extracting the sequential behavior and assessing the potential risk of future violence is not straightforward.

\section{Conclusion}
Domestic violence against women is a worrisome problem that is proven to have a gradual effect of violence, not rarely ending in murder or femicide. Due to the underestimation of such crimes, it is hard to evaluate the level of danger the victims are situated. Therefore, it is important to have tools or mechanisms to assess risk for the women involved in domestic violence, aiming to protect them from worsening the case and prevent fatalities.

Some useful tools have been proposed to understand the level of harm a female victim can be such as the “Violent-o-meter”, which provides a scale of gradual violent behavior from the offender. We proposed a complementary tool, based on a machine learning technique, able to evaluate risk in domestic violence against female victims, thus, having not only a human assessment but also a trained machine interpretation.

Our goal in this work was to provide risk assessment based on police report information using artificial intelligence, or more specifically natural language processing. femicide cases were collected from the Civil Police of the Federal District in Brazil, from 2017 to 2021, along with previous domestic violence preceding these murders. A total of 162 reports were analyzed containing brief written descriptions collected from police officers. We used LSTM due to its good performance with sequentially written data and previous work that demonstrated the potentiality of the algorithm along with NLP (Natural Language Processing) methods. In this work, we followed two methodologies frequently observed with the support of LSTM.

The first approach involved using LSTM purely as a classification method. After preprocessing the text from the 162 reports, we employed LSTM to classify them as either Higher Risk (if the report preceded the femicide by less than one year) or Lower Risk (if the report was filed more than one year prior to the final murder). When tested on 30\% of the data, the algorithm achieved an accuracy of 66\%.

In the second approach, we employed LSTM to predict the next action in a sequence of behaviors observed in the reports. We manually extracted key features from the texts, preprocessed them, and inputted them into the algorithm. These features were identified by observing explicit and implicit content, such as Physical Aggression, Threat, Stalking, Damaging Objects, among others. The results demonstrated the effectiveness of this tool in understanding the sequential behavior in domestic violence crimes and predicting the next action in a series of events.

However, our work has some limitations that require attention. Although the results are promising, it is premature to consider the analysis applicable to real police work, as the data sample used is still relatively small for an artificial intelligence application. Additionally, a deeper examination of bias within the data should be conducted. Furthermore, the manual feature extraction process should be evaluated by other specialists to explore the inclusion or exclusion of additional potential parameters. In future work, we intend to automate the feature extraction process using other Information Retrieval techniques and test alternative algorithms, such as Attention mechanisms \cite{vaswani2017attention}. The challenge lies in developing a tool capable of extracting not only keywords but also implicit and relevant content, which may require advanced machine-learning classification methods. It is worth noting that another limitation is the varying level of detail in the reports. This research also contributes to highlighting the importance of a detailed description of the victim's abuse to provide a confident risk assessment.

In conclusion, this work demonstrates that NLP and LSTM are potential tools for supporting and complementing risk assessment in cases of domestic violence. Through the two studied methods, we were able to gain a better understanding of the level of violence endured by the victims and the potential future consequences if no measures are taken to ensure their safety. Although lives have already been lost to femicide, this research presents an opportunity to leverage the data to study the nuances of such cases more deeply and prevent domestic violence situations from worsening. Machine learning, particularly LSTM, can supplement human evaluations and assist the police and social workers in combating emerging cases.

\section*{Acknowledgment}

We appreciate the support of the Forensic Science non-profit organization Ilaraine Acácio Arce, and the Civil Police of the Federal District, in Brasilia, Brazil, for providing data and investing not only on this but many other important research topics. 

\bibliographystyle{IEEEtran}
\bibliography{feminicide_paper_citations}

\end{document}